\documentclass[preprint,12pt]{article}

\usepackage{graphicx}
\usepackage{amsmath}
\usepackage{amssymb}
\usepackage{booktabs}
\usepackage{comment}
\usepackage[applemac]{inputenc} 
\usepackage{latexsym}
\usepackage{color,graphicx,shortvrb}
\usepackage{amsmath, amssymb}
\usepackage{amsfonts}
\usepackage[colorlinks, bookmarks=true]{hyperref}
\usepackage{algorithm}
\usepackage{algpseudocode}
\usepackage{natbib}

\usepackage{algpseudocode}
\usepackage{amsmath}

\newtheorem{theorem}{Theorem}[section]
\newtheorem{lemma}[theorem]{Lemma}
\newtheorem{remark}[theorem]{Remark}

\usepackage{amssymb}

\begin{document}




\title{Fast Normalized Cross-Correlation for Template Matching with Rotations}


\author{Jos\'e Mar\'ia Almira\textsuperscript{1}, Harold Phelippeau\textsuperscript{2}, \\
Antonio Martinez-Sanchez\textsuperscript{3, {*}}}
\maketitle

{\small \noindent \textsuperscript{1}Dep. Engineering and Computers Technology, Applied Mathematics, University of Murcia, Campus de Espinardo, 30100 Murcia, Spain\\
\textsuperscript{2}Materials and Structural Analysis Division, Advanced Technology, Thermo Fisher Scientific, Bordeaux, France\\
\textsuperscript{3}Dep. Information and Communications Engineering, University of Murcia, Campus de Espinardo, 30100 Murcia, Spain\\
\textsuperscript{{*}}Corresponding author: anmartinezs@um.es}


\begin{abstract}
   {\small Normalized cross-correlation is the reference approach to carry out template matching on images. When it is computed in Fourier space, it can handle efficiently template translations but it cannot do so with template rotations. Including rotations requires sampling the whole space of rotations, repeating the computation of the correlation each time. 
    
    This article develops an alternative mathematical theory to handle efficiently, at the same time, rotations and translations. Our proposal has a reduced computational complexity because it does not require to repeatedly sample the space of rotations. To do so, we integrate the information relative to all rotated versions of the template into a unique symmetric tensor template -which is computed only once per template-. Afterward, we demonstrate that the correlation between the image to be processed with the independent tensor components of the tensorial template contains enough information to recover template instance positions and rotations.
    
    Our proposed method has the potential to speed up conventional template matching computations by a factor of several magnitude orders for the case of 3D images.}
\end{abstract}
%
%







\section{Introduction}\label{sec:intro}

A classical problem in image processing and, particularly, in pattern recognition, is to identify if a large image contains copies -and how many, and their locations and orientations- of a small image, named ``template''. The resulting algorithms are generically known as template matching algorithms \citep{Brun2009, ForPonc2002, GONZWOODS}. The most classical solution is based on using cross-correlations, although there are other approaches based, for example, in metaheuristic algorithms \citep{RePEc:eee:matcom:v:206:y:2023:i:c:p:130-146} or on deep learning \citep{Brun2009, Lamm2022, Moebel2021}. In this paper, we show the mathematical foundations of the cross-correlation-based template matching algorithm (TM in all that follows), and we introduce a new fast algorithm that solves the problem using tensors.

The main advantages of TM, when compared to the algorithms based on machine learning, are that TM is a white box model, it is directly applicable when you have just one template and one larger image (not requiring any kind of training, which may be a very difficult task in some applications), and locates rotations with arbitrary precision (Current deep learning based algorithms for template matching in three-dimensional images are not able to estimate rotations accurately \citep{Lamm2022}).

On the other hand, a major drawback of TM is its computational cost. TM basic idea is to compute the inner product between the (rotated) template and the (translated) image, and normalize the result. These computations are made, for each rotation, in the Fourier domain to efficiently address translations \citep{Lewis1995, Bohm2000, Roseman2003}. However, this process has to be repeated for every rotation to be investigated, thus the resulting complexity has a dependency with the rotations processed. The computational cost of this process may become restrictive for 3D images since $SO(3)$, the space of rotations of $\mathbb{R}^3$, is a (compact) manifold of dimension $3$. In application domains such as cryo-electron microscopy, there are required more than ten thousand rotations for achieving an angular precision of a few degrees. 
 
We propose an algorithm called tensorial template matching, TTM, which integrates into a unique symmetric tensor the information relative to the template in all rotations. In other words, the tensor template incorporates in a unique object the information about all rotations of the template, thus allowing us to find the position and rotation of instances of the template in any tomogram with just a few correlations with the linearly independent components of the tensor. The tensor template is computed only once per template, and, as soon as it is generated, it enables to process any image.
    
\section{Classical template matching}\label{match}

Let us introduce some notation. $d$-dimensional images are just elements of $L^2(\mathbb{R}^d)$, which is a Hilbert space with the inner product $\langle f,g\rangle=\int_{\mathbb{R}^d}f(x)g(x)dx$. It is natural to use the inner product to compare two images $f,g$ of the same size. Concretely, we can use that $\langle f,g\rangle =\|f\|_2\|g\|_2\cos\theta$,
 where $\theta$ is the angle formed by $f$ and $g$. In particular, $f=\alpha g$ for some positive constant $\alpha$ if $\frac{\langle f,g\rangle}{\|f\|_2\|g\|_2}=1$. 
 
Template matching is typically used to study if instances of a ``small'' image $t$ (the template) is present in a larger image $f$, e.g. look for instances of an specific macromolecule in a cryo-electron tomogram (3D volumetric image). The size of the image is connected to the set of points where the image does not vanish, the support of the image. That is, $t$ is meant ``small'' when the set $K=supp(t)=\overline{\{x:t(x)\neq 0\}}^{\mathbb{R}^d}$ is small (e.g., is a subset of a small ball $\mathbb{D}$). 
Let's assume that $f$ and $t$ have quite different sizes, so our interest is to compare $t$ (the template, the small image) 
with just a part of $f$. In such case we need to introduce some special operators $S:L^2(\mathbb{R}^d)\to L^2(\mathbb{R}^d)$ that fix our attention in just a part of the domain of $f$. An interesting example of such operators is
\begin{equation}\label{restriccion}
S_r(f)(x)=U(1-\frac{1}{r}\|x\|_2)f(x)=\left\{ \begin{array}{llll} f(x) & ,\ \|x\|_2\leq r  \\ 0 &, \text{ otherwise} \end{array} \right. ,
\end{equation}
where $U:\mathbb{R}\to\mathbb{R}$ denotes Heaviside's unit step function and $r>0$.   If the support of the template $t$ is $\mathbb{D}_{\mathbf{0}}(r)=\{x:\|x\|_2\leq r\}$, the ball of radius $r$ centered at $\mathbf{0}\in\mathbb{R}^d$, the normalized inner product 
$$\frac{\langle S_r(f),t\rangle}{\|S_r(f)\|_2\|t\|_2}$$ 
informs about the similarity between $t$ and the restriction of $f$ to $\mathbb{D}_{\mathbf{0}}(r)$.  Moreover, if we introduce the translation operator $\tau_{x}: L^2(\mathbb{R}^d)\to L^2(\mathbb{R}^d)$, $\tau_{x}(f)(z)=f(z+x)$ and compute   $$\frac{\langle S_r(\tau_{x}(f)),t\rangle}{\|S_r(\tau_{x}(f))\|_2\|t\|_2}$$  the result informs about the similarity between $t$ and the restriction of $f$ to $\mathbb{D}_{x}(r)=\{z:\|x-z\|_2\leq r\}$. Of course, it may happen that $f$ contains a copy of a rotated version of $t$, so rotations are also necessary for a complete discussion of the problem. Thus, given $R\in SO(d)$, we define the operator $O_R: L^2(\mathbb{R}^d)\to L^2(\mathbb{R}^d)$, $O_R(t)(z)=t(Rz)$, and for $t\in L^2(\mathbb{R}^d)$, we define a rotated version of $t$, 
\begin{equation}
    t_R=O_{R^{-1}}(t).
\end{equation}
 The normalized inner product 
 $$\frac{\langle S_r(\tau_{x}(f)),t_R\rangle}{\|S_r(\tau_{x}(f))\|_2\|t\|_2}$$ 
 informs about the similarity of $t_R$ and the restriction of $f$ to $\mathbb{D}_{x}(r)$. It is important to notice that $\|t\|_2=\|t_R\|_2$.

The operator $S_r$ defined by \eqref{restriccion} has some special properties. Concretely, it is symmetric, semidefinite positive and commutes with rotations 
Recall that, given $(X,\langle\cdot,\cdot\rangle_X)$ a (real) Hilbert space\footnote{In all that follows, we will also use the notation $x\cdot y$ to denote $\langle x,y\rangle$, if this simplifies computations.}, an operator $S:X\to X$ is named: 
\begin{itemize}
\item Symmetric (also named self-adjoint) if
\[
\langle f,S(g)\rangle_X= \langle S(f),g\rangle_X \text{ for all } f,g\in X
\] 
\item Semidefinite positive, if 
\[
\langle f,S(f)\rangle_X\geq 0  \text{ for all } f\in X
\] 
\item Definite positive, if it is semidefinite positive and $\langle f,S(f)\rangle =0$ implies $f=0$.
\end{itemize}
If $S:X\to X$ is a symmetric semidefinite positive operator (SSP, in all what follows), then $X$ becomes a pre-Hilbert space with the inner product 
\begin{equation} \label{product}
    \langle f,g\rangle_S:= \langle f,S(g)\rangle_X
\end{equation}
and the seminorm 
\begin{equation} \label{seminorm}
    \|f\|_S=\sqrt{\langle f,f\rangle_S}
\end{equation}
Observe, for example, that if $S$ is given by \eqref{restriccion}, then $\|f\|_S=0$ means that $f_{|\mathbb{D}_{\mathbf{0}}(r)}=0$ almost everywhere. 
\begin{theorem}\label{similar}
Set $X=L^2(\mathbb{R}^d)$. Let $S:L^2(\mathbb{R}^d)\to L^2(\mathbb{R}^d)$ be an SSP operator, and consider the inner product given by \eqref{product}. 
Then 
\begin{itemize}
\item[$(a)$] $ \langle f,g\rangle_S\leq \left|\langle f,g\rangle_S\right| \leq \|f\|_S\|g\|_S$ for all $f,g\in L^2(\mathbb{R}^d)$.
\end{itemize}
Moreover, if $f,g\in L^2(\mathbb{R}^d)$, $\|g\|_S\neq 0$, the following are equivalent statements:
\begin{itemize}
\item[$(b)$] $\langle f,g\rangle_S=\|f\|_S\|g\|_S$.
\item[$(c)$]  $\| f-\frac{\|f\|_S}{\|g\|_S}g\|_S=0$.
\end{itemize}
\end{theorem} 
\noindent \textbf{Proof.} As $S$ is SSP, we have that, for all $\alpha\in\mathbb{R}$, 
\begin{equation}\label{uno}
0\leq \langle f+\alpha g,f+\alpha g\rangle_S=\|f\|_S^2+2\alpha\langle f,g\rangle_S+\alpha^2\|g\|_S^2.
\end{equation}
Hence, if $\|g\|_S\neq 0$, the only way that the quadratic polynomial (in $\alpha$) above is nonnegative everywhere is that 
\[
4\langle f,g\rangle_S^2-4\|f\|_S^2\|g\|_S^2\leq 0,
\]
which is equivalent to 
\begin{equation}\label{dos}
\left|\langle f,g\rangle_S\right|\leq \|f\|_S\|g\|_S. 
\end{equation}
On the other hand, if $\|g\|_S=0$, the only way to satisfy \eqref{uno} is $\langle f,g\rangle_S=0$, in whose case \eqref{dos} also holds. This proves $(a)$. 

Let us now demonstrate $(b)\Leftrightarrow (c)$ whenever $\|g\|_S\neq 0$. Indeed, $(c)$ is equivalent to
\begin{eqnarray*}
0 &=&\| f-\frac{\|f\|_S}{\|g\|_S}g\|_S^2\\ 
&=& \langle f-\frac{\|f\|_S}{\|g\|_S}g, f-\frac{\|f\|_S}{\|g\|_S}g\rangle \\
&=& \|f\|_S^2+ \frac{\|f\|^2_S}{\|g\|^2_S}\|g\|_S^2-2 \frac{\|f\|_S}{\|g\|_S}\langle f,g\rangle_S\\
&=& 2\|f\|_S^2-2 \frac{\|f\|_S}{\|g\|_S}\langle f,g\rangle_S,
\end{eqnarray*} 
which holds if and only if 
\[
\langle f,g\rangle_S=\|f\|_S\|g\|_S.
\]
Thus $(b)\Leftrightarrow (c)$. 
{\hfill $\Box$}

Note that, if $f,t\in L^2(\mathbb{R}^d)$ are two images, $\alpha>0$, and we take $S=S_r$ given by \eqref{restriccion}, 
then $\|\tau_xf-\alpha O_{R^{-1}}(t)\|_S=0$ means that $f$ has a match with $t_R$ in the unit ball centered at $x$. Indeed, there are many ways to define operators $S$ with the property that $\|f-g\|_S=0$ means that $f=g$ in a neighbourhood of $\mathbf{0}$, so that $\|\tau_xf-\alpha O_{R^{-1}}(t)\|_S=0$ means that $f$ has a match with a rotated version of $t$ in a neighbourhood of $\mathbf{0}$. Although arbitrary SSP operators may not enjoy this property, they allow the creation of a general way to deal with this kind of operators.

Thus, in all what follows, we assume that  $S:L^2(\mathbb{R}^d)\to L^2(\mathbb{R}^d)$ is an SSP operator and that $\|\mathbf{1}\|_S^2=\langle \mathbf{1},\mathbf{1}\rangle_S>0$, where $\mathbf{1}(x)=1$ is the constant image.\footnote{$\mathbf{1}$ is not an element of $L^2(\mathbb{R}^d)$, but this can be managed in several ways. In fact, although we use $L^2(\mathbb{R}^d)$ to denote the space of $d$-dimensional images, in practice we only consider images $f$ with compact support $K$. Then, when we compute $\langle f,\mathbf{1}\rangle$, we mean $\langle f,\mathbf{1}\rangle = \int_{\mathbb{R}^d}f(x)dx= \int_{K}f(x)dx=\langle f,\mathbf{1}\chi_K\rangle$. Moreover, since our interest is on operators $S$ that vanish on functions vanishing outside of a certain neighbourhood $\mathbf{D}$ of $0$, by  $\langle \mathbf{1},\mathbf{1}\rangle_S$ we mean $\langle \mathbf{1}\chi_{\mathbf{D}},\mathbf{1}\chi_{\mathbf{D}}\rangle_S$. } 

Rotations and composition of operators will play an important role in this paper. Thus, it is natural to ask how the composition of rotations acts on the images. This is, indeed, a simple computation: 

\begin{equation*}
\begin{split}
O_{R_1R_2}(t)(z)=t(R_1R_2z)
=t(R_1(R_2z))\\
=O_{R_1}(t)(R_2z)=O_{R_2}(O_{R_1}(t))(z)
\end{split}
\end{equation*}
Hence
\begin{equation}   O_{R_1R_2}=O_{R_2}\circ O_{R_1}
\end{equation}
and
\begin{equation}
\begin{split}
t_{R_1R_2}=O_{(R_1R_2)^{-1}}(t)= O_{R_2^{-1}R_1^{-1}}(t) \\
=O_{R_1^{-1}}\circ O_{R_2^{-1}}(t) =(t_{R_2})_{R_1}.
\end{split}
\end{equation}

Given an image $f$, we consider its projection onto the space of images which are $S-$ orthogonal to the constant image $\mathbf{1}$,
\begin{equation}
    P_S(f)=f-\frac{\langle f,\mathbf{1}\rangle_S}{\langle \mathbf{1},\mathbf{1}\rangle_S} \mathbf{1}.
\end{equation}
\begin{remark}
   These projections are important to study invariant properties with respect to constant brightness changes in the images of translations and rotations. Note that there is no ``real'' difference between an image $f$ and the images of the form $f+\alpha \mathbf{1}$, $\alpha\in\mathbb{R}$. When we modify the constant $\alpha$, what we observe is a uniform change in the density or the brightness, but not the apparition of new structures or forms, in the image $f$. Thus, $f$ and its projection $P_S(f)$ essentially represent the very same image, since $f=P_S(f)+\alpha \mathbf{1}$ for certain $\alpha\in\mathbb{R}$.  
\end{remark}
 
Given two images $f,t$, we have that
\[
f=P_S(f)+\alpha \mathbf{1} \text{ and } t=P_S(t)+\beta \mathbf{1} \text{ for certain constants } \alpha,\beta.
\]
Hence
\begin{eqnarray*}
\langle f,t\rangle_S &=& \langle P_S(f)+\alpha \mathbf{1} ,P_S(t)+\beta \mathbf{1}\rangle_S\\
&=&  \langle P_S(f) ,P_S(t)\rangle_S+  \alpha \beta \langle  \mathbf{1} ,\mathbf{1}\rangle_S
\end{eqnarray*}
since $P_S(f),P_S(t) \perp_S \mathbf{1}$. Consequently, if $x\in \mathbb{R}^d$ and $R\in SO(d)$, there are two constants $\rho=\rho(x)$ and $\delta=\delta(R)$ such that 
\begin{eqnarray*}
\langle \tau_x(f),t_R\rangle_S &=& \langle P_S(\tau_x(f)) ,P_S(t_R)\rangle_S+  \rho \delta \langle  \mathbf{1} ,\mathbf{1}\rangle_S.
\end{eqnarray*}
Assume that $S$ commutes with rotations, and take $x\in \mathbb{R}^d$ fixed. Then, for each $R\in SO(d)$ we have that 
\begin{equation}\label{conmuta}
\begin{split}
t_R=O_{R^{-1}}(t)=O_{R^{-1}}(P_S(t)+\beta \mathbf{1})\\
=O_{R^{-1}}(P_S(t))+\beta \mathbf{1}
\end{split}
\end{equation}
since $O_{R^{-1}}(\mathbf{1})=\mathbf{1}$.  Moreover
\begin{eqnarray*}
&\ & \langle O_{R^{-1}}(P_S(t)),\mathbf{1}\rangle_S \\
&=& \int_{\mathbb{R}^d}P_S(R^{-1}u)S(\mathbf{1})(u)du \\
&=&  \int_{\mathbb{R}^d}P_S(v)S(\mathbf{1})(Rv)dv 
\end{eqnarray*}
(just take  $v=R^{-1}u$ and use that $\det R=1$ )
\begin{eqnarray*}
&=&  \int_{\mathbb{R}^d}P_S(t)(v)(O_R\circ S)(\mathbf{1})(v)dv\\
&=&  \int_{\mathbb{R}^d}P_S(t)(v)(S\circ O_R)(\mathbf{1})(v)dv 
\end{eqnarray*}
(since $O_R\circ S= S\circ O_R$)
\begin{eqnarray*}
&=&  \int_{\mathbb{R}^d}P_S(t)(v)(S)(\mathbf{1})(v)dv 
\end{eqnarray*}
(since $O_R(\mathbf{1})=\mathbf{1}$ )
\begin{eqnarray*}
&=& \langle P_S(t),\mathbf{1}\rangle_S = 0.
\end{eqnarray*}
Hence $O_{R^{-1}}(P_S(t))\perp_S \mathbf{1}$ and this, in conjunction with \eqref{conmuta}, implies that  
$$P_S(t_R)=P_S(O_{R^{-1}}(t))= O_{R^{-1}}(P_S(t))=P_S(t)_R.$$
Hence
\[
t_R=P_S(t)_R+\beta \mathbf{1}
\]
is an $S$-orthogonal decomposition of $t_R$, which means that the constant $\beta$ that multiplies $\mathbf{1}$ in the $S$-orthogonal decomposition of $t_R$ does not depend on $R$,  and 
\begin{eqnarray*}
\langle \tau_x(f),t_R\rangle_S &=& \langle P_S(\tau_x(f)) ,P_S(t)_R\rangle_S+  \rho \beta \langle  \mathbf{1} ,\mathbf{1}\rangle_S\\ 
&=& \langle P_S(\tau_x(f)) ,P_S(t_R)\rangle_S+  \rho \beta \langle  \mathbf{1} ,\mathbf{1}\rangle_S.
\end{eqnarray*}
In  particular, for each $x\in \mathbb{R}^d$, the problems: 
\begin{itemize}
\item Maximize $\langle \tau_x(f),t_R\rangle_S $ over rotations $R$.
\item Maximize $\langle P_S(\tau_x(f)) ,P_S(t)_R\rangle_S $ over rotations $R$. 
\item Maximize $\langle P_S(\tau_x(f)) ,P_S(t_R)\rangle_S $ over rotations $R$. 
\end{itemize}
are equivalent.

Let us define:
\begin{equation}
    f_{-x,R^{-1}}:=(O_R\circ \tau_x)(f).
\end{equation}

\begin{lemma} If $S$ is an SSP operator that commutes with rotations, the parameter $\delta$ that appears in the $S$-orthogonal decomposition 
\[
f_{-x,R^{-1}}=P_S(f_{-x,R^{-1}})+\delta \mathbf{1}
\]
does not depend on $R$. Consequently, given $x\in \mathbb{R}^d$, the problems
\begin{itemize}
\item Maximize $\langle f_{-x,R^{-1}},t \rangle_S $ over rotations $R$.
\item Maximize $\langle P_S(f_{-x,R^{-1}}) ,P_S(t)\rangle_S $ over rotations $R$. 
\end{itemize}
are equivalent. 
\end{lemma}

\noindent \textbf{Proof.}  
We know that $\delta=\frac{\langle f_{-x,R^{-1}},\mathbf{1} \rangle_S}{\langle \mathbf{1},\mathbf{1} \rangle_S}$, so that we only need to prove that $\langle f_{-x,R^{-1}},\mathbf{1} \rangle_S$ does not depend on $R$. Indeed, 
\begin{eqnarray*}
\langle f_{-x,R^{-1}},\mathbf{1} \rangle_S &=& \int_{\mathbb{R}^d}f(Ry+x)S(\mathbf{1})(y)dy 
\end{eqnarray*}
(Make the change of variable $z=Ry$ )
\begin{eqnarray*}
&=& \int_{\mathbb{R}^d}f(z+x)S(\mathbf{1})(R^{-1}z)dz \\
&=& \int_{\mathbb{R}^d}f(z+x)(O_{R^{-1}}\circ S)(\mathbf{1})(z)dz \\
&=& \int_{\mathbb{R}^d}f(z+x)(S\circ O_{R^{-1}})(\mathbf{1})(z)dz 
\end{eqnarray*}
(since $S$ commutes with $O_{R^{-1}}$ )
\begin{eqnarray*}
&=& \int_{\mathbb{R}^d}f(z+x)S(\mathbf{1})(z)dz 
\end{eqnarray*}
(since $O_{R^{-1}}(\mathbf{1})=\mathbf{1}$ )
\begin{eqnarray*}
&=& \langle \tau_x(f),\mathbf{1}\rangle_S
\end{eqnarray*}
{\hfill $\Box$}

We can now state and demonstrate the following:
\begin{theorem}[Classical template matching] \label{ThCTM}Let $S$ be a SSP operator which commutes with rotations and let $x\in \mathbb{R}^d$ be fixed. Then the following are equivalent problems:
\begin{itemize}
\item[$(a)$] Maximize $\langle f_{-x,R^{-1}},t \rangle_S$ over rotations $R$.
\item[$(b)$] Maximize $\langle \tau_x(f) , t_R \rangle_S$ over rotations $R$.
\item[$(c)$] Maximize $\langle P_S(\tau_x(f)), P_S(t_R) \rangle_S$ over rotations $R$.
\item[$(d)$] Maximize $\langle P_S(f_{-x,R^{-1}}),P_S(t) \rangle_S $ over rotations $R$.
\end{itemize}
Moreover, if $\|t\|_S>0$ and $S$ also has the property that $\|f\|_S=0$ implies $f_{|\mathbf{D}}=0$ for a certain neighborhood $\mathbf{D}$ of $\mathbf{0}\in\mathbb{R}^d$ which contains the supports of all the rotated templates $t_Q$ with $Q\in SO(d)$, then a match between $f$ and $t_R$ in $x$ is got whenever any one of the following claims hold:
\begin{itemize}
\item[$(a^*)$] $\frac{\langle f_{-x,R^{-1}},t \rangle_S}{\| f_{-x,R^{-1}}\|_S\|t\|_S}=1$  
\item[$(b^*)$] $\frac{\langle \tau_x(f) , t_R \rangle_S}{\| \tau_x(f)\|_S\|t_R\|_S}=1$
\item[$(c^*)$] $\frac{\langle P_S(\tau_x(f)), P_S(t_R)  \rangle_S}{\| P_S(\tau_x(f))\|_S \|P_S(t_R)\|_S} =1$ 
\item[$(d^*)$] $\frac{\langle P_S(f_{-x,R^{-1}}),P_S(t) \rangle_S}{\|P_S( f_{-x,R^{-1}})\|_S\|P_S(t)\|_S}=1$ 
\end{itemize}
Finally, the normalized correlations described in $(a^*)$, $(b^*)$, $(c^*)$, and $(d^*)$ do not change when we substitute $f$ by $\alpha f+\beta$, and $t$ by $\delta t+\gamma$, with $\alpha,\beta,\delta,\gamma \in\mathbb{R}$, $\alpha,\delta \neq 0$. 
\end{theorem}

\noindent \textbf{Proof.}   The equivalences $(a)\Leftrightarrow (d)$ and $(b)\Leftrightarrow (c)$ have been already shown. The following identities demonstrate $(a)\Leftrightarrow (b)$:
\begin{eqnarray*}
\langle f_{-x,R^{-1}},t\rangle_S &=& \int_{\mathbb{R}^d}f(Rz+x)S(t)(z)dz\\
&=& \int_{\mathbb{R}^d}f(y+x)S(t)(R^{-1}y)dy 
\end{eqnarray*}
( just take $Rz=y$ and use that $\det R=1$ )
\begin{eqnarray*}
&=& \int_{\mathbb{R}^d}f(y+x)(O_{R}^{-1}\circ S)(t)(y)dy\\
&=& \int_{\mathbb{R}^d}f(y+x)(S\circ O_{R}^{-1})(t)(y)dy  
\end{eqnarray*}
(since $S$   commutes with $O_{R^{-1}}$ )
\begin{eqnarray*}
&=& \langle \tau_x(f),t_R\rangle_S \\
\end{eqnarray*}
The other claims are a direct consequence of Theorem \ref{similar}.

{\hfill $\Box$}

In all that follows, we assume that $S$ is an SSP operator that commutes with rotations and $t$ is normalized in the sense that $t\perp_S \mathbf{1}$ and $\|t\|_S=1$. Then $P_S(t_R)=t_R$ and $\|t_R\|_S=1$ for all rotation $R$. Consequently, 
\begin{equation*}
     \langle \tau_x(f),t_R\rangle_S = \langle P_S(\tau_x(f))+\alpha \mathbf{1},t_R\rangle_S \\
    = \langle P_S(\tau_x(f)),t_R\rangle_S 
    =  \langle P_S(\tau_x(f)),P_S(t_R)\rangle_S 
\end{equation*}
and 
\begin{equation*}
    c(x,R) =\frac{\langle P_S(\tau_x(f)),P_S(t_R)\rangle_S}{\|P_S(\tau_x(f))\|_S \|P_S(t_R)\|_S} \\ = \frac{\langle \tau_x(f),t_R\rangle_S}{\|P_S(\tau_x(f))\|_S } 
\end{equation*}
attains its maximum ($=1$) if and only if there is a perfect match between $f$ and $t_R$ in $x$. Moreover, if we define $w(x)=\frac{1}{\|P_S(\tau_x(f))\|_S }$ and consider the cross-correlation of functions $f,g\in L^2(\mathbb{R}^d)$, which is defined by
\begin{equation}\label{correlation}
(f\star g )(x)=\int_{\mathbb{R}^d}f(z+x)g(z)dz =\langle \tau_x(f),g\rangle,
\end{equation}
then 
\begin{equation}
    c(x,R) = w(x) (f\star S(t)_R)(x).
\end{equation}

 A perfect match is, in general terms, never attained. This is so because the desired image, represented by the template $t$, is usually supported on a strict subset $\Omega$ of the domain $D$ were the operator $S$ is able to distinguish functions. Thus, the image $f$ may well contain a copy of the image represented by $t_R$ but in the neighbourhoods of the support of $t_R$, $f$ will contain some information which is not present in $t_R$. In addition, $f$ is usually corrupted by noise and distortions. This means that the normalized correlations described in items $(a^*)-(d^*)$ of Theorem \ref{ThCTM}, will never equal $1$. Consequently, a threshold should be introduced in order to decide if a match has (or has not) been produced.  

In order to find the rotation which maximizes $c(x,R)$, the cross-correlation $(f\star S(t)_R)(x)$ should be computed for a huge amount of rotations $R$, which makes classical matching an inefficient approach for template matching. Indeed, for $d=3$, the size of the set of rotations $R$ used to sample $SO(3)$ well enough to guarantee a reliable result varies between $10^4$ and $5\cdot 10^5$ rotations \citep{Chaillet2023}.

     Due to numerical reasons, high frequencies may be altered during rotation transformation. Thus, in practice, we do not apply the operator $S$ to the original images $f,t$ but to a filtered version of them that eliminates these high frequencies. Concretely, we apply an isotropic (i.e. rotation invariant) low-pass filter $h$ to both images and, after that, we apply the template matching algorithm to the resulting images. The idea behind this is that, if there is a match between $f$ and $t$, there will be a match between $\mathfrak{f}=f\ast h$ and $\mathfrak{t}=t\ast h$ too. The operator $S$ results from applying a rotationally symmetric mask $m(x)=\rho(\|x\|)$ to the given image. Thus, we substitute $f$ by $\mathfrak{f}=f\ast h$ and $t$ by $\mathfrak{t}=t\ast h$. Then we apply the classical (or tensor) matching algorithm to the pair of images $\mathfrak{f},\mathfrak{t}$ using the SSP operator $S(\mathfrak{f})(x)=m(x) \mathfrak{f}(x)$. Usually, the mask $m$ equals $1$ within a certain radius around $\mathbf{0}$ and equals $0$ outside a sightly larger radius. In between these radii the mask takes values between 0 and 1. Under these restrictions, it is clear that the operator $S$ is SSP and commutes with rotations. Moreover, if $0=\|f\|_S=\langle f,S(f)\rangle \geq \int_{\mathbf{D}}f^2(x)dx\geq 0$, we have that $f_{|\mathbf{D}}=0$ where $\mathbf{D}$ is a ball of positive radius centered at $\mathbf{0}$. 
Let us compute the inner product 
     \begin{eqnarray*}
         \langle \tau_x(\mathfrak{f}),\mathfrak{t}_R\rangle_S &=& \langle \tau_x(\mathfrak{f}),m\mathfrak{t}_R\rangle\\
         &=& \langle \tau_x(f\ast h),m(t\ast h)_R\rangle \\
         &=& \langle \tau_x(f)\ast h,m(t_R\ast h)\rangle 
          \end{eqnarray*}
         (since every filter is translation invariant, and $h$ is isotropic)
          \begin{eqnarray*}
          &=& \langle \tau_x(f),h \ast (m(t_R\ast h))\rangle 
          \end{eqnarray*}
        (use $\widetilde{h}(x):=h(-x)=h(x)$, which follows from isotropy of  $h$ )
    \begin{eqnarray*}
          &=& \langle \tau_x(f),t_R\rangle_{\overline{S}} 
    \end{eqnarray*}
     where $$\overline{S}(f)=h \ast (m\cdot(f\ast h))$$
and we use $\cdot$ to denote the standard product of real functions.       
     This means that we would have the same effect just considering the template matching algorithm associated with the operator $\overline{S}$ applied to the images $f,t$. Moreover, the following holds: 
     \begin{lemma}\label{opSSP}
     Let $S:L^{2}(\mathbb{R}^d)\to L^{2}(\mathbb{R}^d)$ be given by 
     \begin{equation} \label{operadorSSP}
         S(f)=h \ast (m\cdot (f\ast h))
     \end{equation}
      with $h$ defining an isotropic filter and $m$ a rotationally symmetric mask as described above. Then $S$ is SSP. 
     \end{lemma}

     \noindent \textbf{Proof.}  
         For the proof, we use the following (well-known) formulae: For functions $a,b,c\in L^{2}(\mathbb{R}^d)$, we have that 
         $(a\star b)(x)=\langle \tau_x(a),b\rangle$, so that $(a\star b)(0)=\langle a,b\rangle = (a\ast \widetilde{b})(0)$, $a\star b=a\ast \widetilde{b}$, and $a\star (b\ast c) = (a\star b)\star c$.

         Let us now consider the product $\langle f,S(f)\rangle $:
         \begin{eqnarray*}
             \langle f,S(f)\rangle &=& (f\star S(f))(0) \\
             &=& (f\star (h \ast (m\cdot(f\ast h))))(0) \\
             &=& ((f\star h) \star (m\cdot(f\ast h)))(0) \\
             &=& \langle f\star h, m\cdot(f\ast h)\rangle \\
             &=& \langle f\ast h, m\cdot(f\ast h)\rangle \geq 0 
         \end{eqnarray*}
         (since $h=\widetilde{h}$ and $m\geq 0$). 
         This proves that $S$ is semidefinite positive. Let us show the symmetry:
         \begin{eqnarray*}
             \langle f,S(g)\rangle &=& (f\star S(g))(0) = (f\star (h \ast (m\cdot(g\ast h))))(0) \\
             &=& ((f\star h) \star (m\cdot(g\ast h)))(0) \\ 
             &=& \langle f\star h, m\cdot(g\ast h)\rangle \\
             &=& \langle m\cdot (f\star h), (h\ast g)\rangle 
             \end{eqnarray*}
        (since $g\ast h= h\ast g $  and $\cdot$ is the standard product of functions)
             \begin{eqnarray*}
             &=& ((m\cdot (f\star h))\star(h\ast g))(0) \\
              &=& (((m\cdot (f\star h))\star h)\star g)(0) \\
              &=& \langle ((m\cdot (f\star h))\star h) , g\rangle \\
              &=& \langle ((m\cdot (f\ast h))\ast h) , g\rangle \text{ (since } h=\widetilde{h} \text{ )}\\
              &=& \langle S(f),g\rangle .
         \end{eqnarray*}
     {\hfill $\Box$} 

In all that follows, we assume that the SSP operator $S$ is of the form \eqref{operadorSSP} with $h$, $m$ verifying the hypotheses of Lemma \ref{opSSP}. Thus, the template matching algorithm is applied with this operator and a fast computation of $c(x,R)$ is needed.

A direct computation leads to:
\begin{eqnarray*}
     c(x,R)  
    &=&  \frac{\langle \tau_x(f),t_R\rangle_S}{\|P_S(\tau_x(f))\|_S }\\
    &=&  \frac{1}{\|P_S(\tau_x(f))\|_S } \langle \tau_x(f),S(t_R)\rangle\\
    &=&  \frac{1}{\|P_S(\tau_x(f))\|_S } ( \tau_x(f)\star S(t_R))(0)\\
    &=&  \frac{1}{\|P_S(\tau_x(f))\|_S } ( \tau_x(f)\star (h\ast (m\cdot (h\ast t_R))))(0)\\
    &=&  \frac{1}{\|P_S(\tau_x(f))\|_S } ( (\tau_x(f)\star h)\star (m\cdot (h\ast t_R)))(0)\\
    &=&  \frac{1}{\|P_S(\tau_x(f))\|_S } ( (\tau_x(f)\ast h)\star (m\cdot (h\ast t_R)))(0) 
    \end{eqnarray*}
 (since $h=\widetilde{h}$)
    \begin{eqnarray*}
    &=&  \frac{1}{\|P_S(\tau_x(f))\|_S } \langle (\tau_x(f)\ast h), m\cdot (h\ast t_R)\rangle \\
    &=&  \frac{1}{\|P_S(\tau_x(f))\|_S } \langle (\tau_x(f)\ast h), (m\cdot (h\ast t))_R\rangle 
\end{eqnarray*}
(since $h$ is isotropic, and $m$ is rotationally symmetric).  Moreover, 
\begin{eqnarray*}
   \|P_S(\tau_x(f))\|_S^2  
   &=&  \|\tau_x(f)-\frac{\langle \tau_x(f),\mathbf{1}\rangle_S}{\|\mathbf{1}\|_S^2}\mathbf{1}\|_S^2\\
   &=&  \|\tau_x(f)\|_S^2
   -2\langle \tau_x(f),\frac{\langle \tau_x(f),\mathbf{1}\rangle_S}{\|\mathbf{1}\|_S^2}\mathbf{1}\rangle_S + \|\frac{\langle \tau_x(f),\mathbf{1}\rangle_S}{\|\mathbf{1}\|_S^2}1 \|_S^2 \\
   &=&  \|\tau_x(f)\|_S^2-2\langle \tau_x(f),\frac{\langle \tau_x(f),S(\mathbf{1})\rangle}{\|\mathbf{1}\|_S^2}S(\mathbf{1})\rangle  + \|\frac{\langle \tau_x(f),S(\mathbf{1})\rangle}{\|\mathbf{1}\|_S^2}\mathbf{1} \|_S^2 \\
   &=&  \|\tau_x(f)\|_S^2 -2\frac{(\langle \tau_x(f),S(\mathbf{1})\rangle)^2}{\|\mathbf{1}\|_S^2} + \frac{(\langle \tau_x(f),S(\mathbf{1})\rangle)^2}{\|\mathbf{1}\|_S^4}\|\mathbf{1} \|_S^2 \\
   &=&  \|\tau_x(f)\|_S^2-\frac{(\langle \tau_x(f),S(\mathbf{1})\rangle)^2}{\|\mathbf{1}\|_S^2}\\
   &=&  \langle \tau_x(f), S(\tau_x(f))\rangle -\frac{(\langle \tau_x(f),S(\mathbf{1})\rangle)^2}{\|\mathbf{1}\|_S^2}
\end{eqnarray*}
Now, using the definition of $S$ (and imposing $h\ast \mathbf{1}=\mathbf{1}$), we can simplify the computation as follows: 
\begin{eqnarray*}
  \|P_S(\tau_x(f))\|_S^2 
  &=&  \langle \tau_x(f), h\ast (m\cdot (h\ast \tau_x(f))\rangle  -\frac{(\langle \tau_x(f),h\ast (m\cdot (h\ast \mathbf{1}))\rangle)^2}{\|\mathbf{1}\|_S^2} \\
   &=&  \langle \tau_x(f)\ast h, m\cdot (\tau_x(f)\ast h)\rangle -\frac{(\langle \tau_x(f)\ast h, m\cdot (h\ast \mathbf{1})\rangle)^2}{\|\mathbf{1}\|_S^2} \\
   &=&  \langle (\tau_x(f)\ast h)^2, m\rangle -\frac{(\langle \tau_x(f)\ast h, m\cdot (h\ast \mathbf{1})\rangle)^2}{\|\mathbf{1}\|_S^2} \\
   &=&  \langle (\tau_x(f)\ast h)^2, m\rangle -\frac{(\langle \tau_x(f)\ast h, m\rangle)^2}{\|\mathbf{1}\|_S^2}
\end{eqnarray*}
Note that the FFT algorithm can be used to compute the inner products appearing at the end of the formula above, which helps to fasten the algorithm. Indeed, if $f,g$ are two images, $\langle f,g\rangle = (f\ast \widetilde{g})(0)$, so that 
\begin{equation}
    \langle f,g\rangle= \mathcal{F}^{-1}( \mathcal{F}(f)\cdot  \mathcal{F}(\widetilde{g}))(0)=  \mathcal{F}^{-1}( \mathcal{F}(f)\cdot \overline{ \mathcal{F}(g)})(0).
\end{equation}
Moreover, the following identities also hold:
\begin{eqnarray*}
    \|\mathbf{1}\|_S^2 &=& \langle \mathbf{1},S(\mathbf{1})\rangle = \langle \mathbf{1},h\ast (m\cdot (h\ast \mathbf{1}))\rangle \\
    &=&  \langle \mathbf{1}\ast h, (m\cdot (h\ast \mathbf{1}))\rangle = \langle \mathbf{1}, m\rangle,\\
    \|P_S(t)\|_S &=&  \sqrt{\langle (h\ast t)^2,m\rangle -\frac{(\langle t\ast h, m\rangle)^2}{\langle \mathbf{1}, m\rangle}},
\end{eqnarray*}
and
\[
\langle t,\mathbf{1}\rangle_S= \langle h\ast t,m\rangle.
\]
Thus,
\begin{eqnarray*}
     m\left(h\ast \frac{P_S(t)}{\|P_S(t)\|_S}\right) 
    &=& m\left(h\ast \frac{t-\frac{\langle t,\mathbf{1}\rangle_S}{\|\mathbf{1}\|_S^2}\mathbf{1}}{\sqrt{\langle (h\ast t)^2,m\rangle -\frac{(\langle t\ast h, m\rangle)^2}{\langle \mathbf{1}, m\rangle}}}\right)\\
    &=& m \frac{h\ast t-\frac{\langle h\ast t,m\rangle}{\langle \mathbf{1}, m\rangle}\mathbf{1}}{\sqrt{\langle (h\ast t)^2,m\rangle -\frac{(\langle t\ast h, m\rangle)^2}{\langle \mathbf{1}, m\rangle}}}
\end{eqnarray*}

    Algorithm \ref{alg:tm}, devoted to classical template matching, has been developed using the formulae above.

An important tool we will use in this paper is the set $\mathbb{H}$ of quaternions. In particular, we will use that rotations can be parametrized by unit quaternions (which can be identified with the unit $3$-sphere $\mathbb{S}^3$), as well as the following formulae (see, e.g. \citep{Ebbetal, Pont}): 
\begin{itemize} 
\item If $x\in \mathbb{H}$ has norm $1$, then $x^{-1}=x^*$. 
\item Given $x\in \mathbb{H}$, $x=a+b\mathbf{i}+c\mathbf{j}+d\mathbf{k}$, we identify $x$ with a pair $(a,v)$ where $a\in\mathbb{R}$ and $v=(b,c,d)\in\mathbb{R}^3$, and call $a$ the real part of $x$, $a=\mathbf{Re}(x)$. Then, if $x=(a,v), y=(b,w)\in \mathbb{H}$, we have that 
\begin{equation*}\label{producto}
    \mathbf{Re} (xy)=ab-\langle v, w \rangle 
\end{equation*}
Consequently, if $x,y\in \mathbb{H}$ have norm $1$, then
\begin{equation}\label{producto2}
\begin{split}   
\langle x,y\rangle = \mathbf{Re}(y^{-1}x) = \mathbf{Re}(xy^{-1}) 
\\ = \mathbf{Re}(yx^{-1})= \mathbf{Re}(x^{-1}y)
\end{split}
\end{equation}
\end{itemize}
We end this section with a well-known result about composition of SSP operators that will be used in the proof of the main theorem of the paper:
\begin{lemma}\label{composition}
If $T,S:L^2(\mathbb{R}^d)\to L^2(\mathbb{R}^d)$ are semidefinite positive symmetric and commute, then $TS$ is also semidefinite positive symmetric.
\end{lemma}

\noindent \textbf{Proof.}  
If $S,T$ are SSP, then they have (unique) positive square roots $\sqrt{T},\sqrt{S}$, which are symmetric operators too. Moreover, the composition $\sqrt{T}\sqrt{S}$ is also symmetric. Since $T,S$ commute, their positive square roots commute. Hence $$TS= \sqrt{T}\sqrt{T}\sqrt{S}\sqrt{S}= \sqrt{T}\sqrt{S}\sqrt{T}\sqrt{S}= (\sqrt{T}\sqrt{S})^2,$$
and the square of any symmetric operator is positive.
{\hfill $\Box$}

\begin{algorithm}
\caption{Classical template matching with rotations}\label{alg:tm}
\begin{algorithmic}
\State \begin{center}  \textbf{Load and Fourier transform of the image }  \end{center}
\State Load data into f
\State $f \gets h\ast f$
\State $\hat{f} \gets \mathcal{F}(f)$
\State $f\gets f^2$
\State $\hat{g} \gets \mathcal{F}(f)$

\State \begin{center}   \textbf{Load mask and pre-process the image  }  \end{center}
\State Load the mask into m
\State $m_{sum} \gets \langle m, 1\rangle $
\State $\hat{m} \gets \mathcal{F}(m)$
\State $\hat{g} \gets \hat{g} \ \overline{\hat{m}}$
\State $w\gets \mathcal{F}^{-1} (\hat{g})$
\State $\hat{m} \gets \hat{f} \ \overline{\hat{m}}$
\State $w_{temp}\gets \mathcal{F}^{-1} (\hat{m})$
\State $w\gets \frac{1}{\sqrt{w-\frac{w_{temp}^2}{m_{sum}}}}$   \ \ \ \ \ \  {\hfill // $w(x)=\frac{1}{\|P_S(\tau_x(f))\|_S}$   }

\State \begin{center}   \textbf{Load and pre-process the template }   \end{center}
\State Load template into t
\State $t\gets h\ast t$
\State $t_1\gets \langle t,m \rangle$
\State $t_2\gets \langle t^2,m \rangle$
\State $t_O\gets \frac{-t_1}{m_{sum}}$
\State $t_S\gets \frac{1}{\sqrt{t_2-t_1^2/m_{sum}}}$
\State $t\gets mt_S(t+t_O)$
\State \begin{center}   \textbf{Compute the output  } \end{center}
\State $t\gets t_R$ \ \ \ \  {\hfill // Rotate $t$ } 
\State $\hat{t}\gets \mathcal{F}(t)$
\State $\hat{t}\gets \hat{f} \overline{\hat{t}}$
\State $c \gets w \mathcal{F}^{-1}(\hat{t})$
\end{algorithmic}
\end{algorithm}

 \section{Tensor template matching}\label{sec:tensor}
This section introduces a tensorial template matching (TTM) algorithm. The purpose is to handle translations and rotations efficiently at the same time. First, we introduce some background necessary to understand further mathematical developments. Second, we present the main theorem for TTM, which allows us to determine the optimal rotation of the template, $t$, on every match in the image $f$ without sampling the $SO(3)$ by computing some tensors. Finally, we explain how to determine match positions (template translations) directly from the computed tensors.

\subsection{Tensor background}
A tensor $A\in T^n(\mathbb{R}^d)$ of order $n$ and dimension $d$ is just an array of the form $A=(A_{i_1,\cdots,i_n})_{1\leq i_1,\cdots,i_n\leq d}$ where all the entries $A_{i_1,\cdots,i_n}$ are real numbers. The tensor $A$ is named symmetric if $A_{i_1,\cdots,i_n}=A_{i_{\sigma(1)},\cdots,i_{\sigma(n)}}$ for every permutation $\sigma \in \Sigma_n$ (the set of permutations of $\{1,\cdots,n\}$). We denote by $S^n(\mathbb{R}^d)$ the set of symmetric tensors of order $n$ and dimension $d$. An important example of symmetric tensor of order $n$ is the so called $n$-th tensor power of a vector $v=(v_1,\cdots,v_d)\in\mathbb{R}^d$, which is defined as
\begin{equation}
    v^{\odot n} =(v_{i_1}v_{i_2}\cdots v_{i_n})_{1\leq i_1,\cdots, i_n\leq d}.
\end{equation}

It is well known that  $T^n(\mathbb{R}^d)$ and $S^n(\mathbb{R}^d)$ are real vector spaces with the natural operations (pointwise sum and multiplication by a scalar), and that $\dim T^n(\mathbb{R}^d)=d^n$, $\dim S^n(\mathbb{R}^d)=\binom{n+d-1}{n}$ for all $n,d\geq 1$. 
For example,  $\dim T^4(\mathbb{R}^4)=4^4=256$, $\dim S^4(\mathbb{R}^4)=\binom{7}{4}=35$. 
Moreover, every symmetric tensor is a finite sum of tensor powers, which allows us to introduce the concept of the (symmetric) rank of a symmetric tensor as the minimal number of tensor powers used to represent the tensor with their sum \citep{Comon}.  

The map $\langle\cdot,\cdot\rangle:  T^n(\mathbb{R}^d)\times   T^n(\mathbb{R}^d) \to\mathbb{R}$ given by 
\begin{equation}
    \langle A, B\rangle = \sum_{i_1=1}^d \sum_{i_2=1}^d \cdots \sum_{i_n=1}^d A_{i_1,\cdots,i_n}B_{i_1,\cdots,i_n}
\end{equation}
defines an inner product. It is also usual to denote $A\cdot B=\langle A,B\rangle$. Moreover, with this notation, if $x,y\in\mathbb{R}^d$ are $d$-dimensional vectors, a direct application of the multinomial theorem shows that 
\begin{equation}
x^{\odot n}\cdot y^{\odot n}=(x \cdot y)^n=(\langle x,y\rangle)^n.
\end{equation}
Moreover, if $A\in S^n(\mathbb{R}^d)$, and $x=(x_1,\cdots,x_d)\in\mathbb{R}^d$, we can also consider the inner product
\begin{equation}
    A\cdot x^{\odot n} =\langle A, x^{\odot n}\rangle = \sum_{i_1=1}^d \sum_{i_2=1}^d \cdots \sum_{i_n=1}^d A_{i_1,\cdots,i_n}x_{i_1} \cdots x_{i_n},
\end{equation}
which can be seen as an homogeneous polynomial in $d$ variables, of degree $n$, which justifies using the notation $Ax^n=A\cdot x^{\odot n}$. Moreover, if $k<n$, $Ax^{k}\in S^{n-k}(\mathbb{R}^d)$ denotes the symmetric tensor whose components are 
\begin{equation}
(Ax^k)_{i_1,\cdots,i_{n-k}} =\\
\sum_{j_1=1}^d \sum_{j_2=1}^d \cdots \sum_{j_k=1}^d A_{i_1,\cdots,i_{n-k},j_1,\cdots,j_k}x_{j_1}\cdots x_{j_k}.
\end{equation}

In particular, $Ax^{n-1}\in S^1(\mathbb{R}^d)=\mathbb{R}^d$ is a vector whose $i$-th component is 
\begin{equation}
(Ax^{n-1})_{i}=  \sum_{j_1=1}^d \sum_{j_2=1}^d \cdots \sum_{j_{n-1}=1}^d A_{i,j_1,\cdots,j_{n-1}}x_{j_1}\cdots x_{j_{n-1}}.
\end{equation}
Indeed, if $\varphi(x)=Ax^n$ then 
\[
\nabla \varphi(x)=nA x^{n-1},
\]
where $\nabla \varphi $ denotes the gradient of the function $\varphi :\mathbb{R}^d\to\mathbb{R}$.  

Note that the vector $x$ can be chosen from $\mathbb{C}^d$ in the definitions above. This justifies the following definition (see \citep{cui}): Given $A\in S^n(\mathbb{R}^d)$, $B\in S^m(\mathbb{R}^d)$. We say that $\lambda\in \mathbb{C}$ is a $B$-eigenvalue of $A$ and $u\in \mathbb{C}^d$ is its associated $B$-eigenvector (equivalently, that $(\lambda,u)$ is a $B$-eigenpair of $A$) if $Au^{n-1}=\lambda Bu^{m-1}$ and $Bu^m=1$. 

Using gradients, we can rewrite the equation $Au^{n-1}=\lambda Bu^{m-1}$ as
\[
\frac{1}{n}\nabla Au^n=\lambda \frac{1}{m}\nabla Bu^m.
\]
Hence $u$ is a $B$-eigenvector of $A$ if and only if it is a critical point of the following optimization problem:
\begin{equation}\label{eigen}
\left\{\begin{array}{llll} \text{Maximize: } & Ax^n\\ \text{under the restriction:} & Bx^m=1.\end{array}\right. 
\end{equation}
Two particularly important cases are the $H$-eigenvectors
 \begin{equation}\label{Heigen}
\left\{\begin{array}{llll} \text{Maximize: } & Ax^n\\ \text{under the restriction:} & \sum_{i=1}^dx_i^m=1\end{array}\right. 
\end{equation}
and the $Z$-eigenvectors
 \begin{equation}\label{Zeigen}
\left\{\begin{array}{llll} \text{Maximize: } & Ax^n\\ \text{under the restriction:} & \sum_{i=1}^dx_i^2=1.\end{array}\right. 
\end{equation}

The optimization problem associated with finding $Z$-eigenvectors of a given symmetric tensor is particularly important for us since the tensor matching algorithm we propose is reduced to one of these problems in each position, and, fortunately, there are good iterative algorithms to approximate the solutions of \eqref{Zeigen} (see e.g.  \citep{Kofidis2001OnTB, Kolda2010ShiftedPM}). These algorithms have a linear rate of convergence. In section \ref{position} we show an heuristics that can be used to select the positions where a match is probable, so that solving \eqref{Zeigen} is necessary.

\subsection{Defining of the Tensor template}

In all that follows in this paper,  $S:L^2(\mathbb{R}^d)\to L^2(\mathbb{R}^d)$ denotes an SSP operator which commutes with rotations and the template $t\in L^2(\mathbb{R}^d)$ is assumed to be normalized by $t\perp_S \mathbf{1}$ and $\|t\|_S=1$.  In section \ref{match} we proved that, for each $x\in\mathbb{R}^d$, $c(x,R) =  w(x) (f\star S(t)_R)(x)$
attains its maximum value on rotation $R$  (and this value equals $1$) if and only if there is a match between $f$ and $t$ at $(x,R)$ (i.e., a match between $\tau_x(f)$ and $t_R$). Let us define the symmetric tensor $C_n(x)\in S^{n}(\mathbb{R}^{d'})$, where $d'$ is the number of parameters used to describe the rotations $SO(d)$ (in particular, for $d=3$, we get $d'=4$) by the formula:
\begin{equation}
    C_n(x)=\int_{SO(d)}R^{\odot n}c(x,R)dR
\end{equation}
This means that 
\begin{equation}
    (C_n(x))_{i_1,\cdots,i_n}=\int_{SO(d)}R_{i_1}R_{i_2}\cdots R_{i_n}c(x,R)dR
\end{equation}
for all $1\leq i_1,\cdots,i_n\leq d'$. Hence
\begin{eqnarray*}
(C_n(x))_{i_1,\cdots,i_n} 
 &=& \int_{SO(d)}R_{i_1}\cdots R_{i_n}c(x,R)dR \\
 &=& w(x)\int_{SO(d)}R_{i_1}\cdots R_{i_n} (f\star S(t)_R)(x)dR \\
 &=& w(x)\int_{SO(d)}R_{i_1}\cdots R_{i_n} \langle \tau_x(f),S(t)_R\rangle dR \\
 &=& w(x)\int_{SO(d)}R_{i_1}\cdots R_{i_n} \int_{\mathbb{R}^d} \tau_x(f)(z)S(t)_R(z)dz dR \\
 &=& w(x)\int_{\mathbb{R}^d} \tau_x(f)(z) \left(\int_{SO(d)}R_{i_1}\cdots R_{i_n}  S(t)_R(z) dR \right)dz\\ 
 &=&  w(x) \langle \tau_x(f),(T(z))_{i_1,\cdots,i_n}\rangle ,
\end{eqnarray*}
where 
\begin{equation}
    T(z)=\int_{SO(d)}R^{\odot n}S(t)_R(z) dR \in S^{n}(\mathbb{R}^{d'})
\end{equation}
is a tensor template (or tensorial needle).

It is of fundamental importance to observe that $T(z)$ is computed only once and contains a reduced number of components, since $\dim S^n(\mathbb{R}^{d'})=\binom{n+d'-1}{n}$ (in particular, $\dim S^4(\mathbb{R}^{4})=\binom{7}{4}=35$). Indeed, this is one of the main reasons why the tensor template matching algorithm we introduce in this paper is fast.  The other reason is that rotations $R$ defining a match between $f$ and $t_R$  at $x$ are $Z$-eigenvectors of the symmetric tensor $C_n(x)$, which is really remarkable because the power method used in \citep{Kolda2010ShiftedPM, Kofidis2001OnTB} for the solution of the corresponding optimization problem does not require using myriads or even millions of rotations -as is the case with classical matching algorithms-  but just a reduced set of them: one by iteration.

\subsection{Finding the correct rotation}
Let us state the main result of this paper:
\begin{theorem}\label{main} 
Let $f,t\in L^2(\mathbb{R}^3)$, $x\in\mathbb{R}^3$, and $n\in 2\mathbb{N}$ be given. If there is a match between $f$ and $t_R$  at $x$, the function $\varphi(Q)=C_n(x)\cdot Q^{\odot n}$, defined on rotations of $\mathbb{R}^3$, when parametrized by unit quaternions $Q$, attains its global maximum at $Q=R$. 
\end{theorem}

\noindent \textbf{Proof}.   We prove the result as a consequence of Theorem \ref{ThCTM}. Thus, our main goal is to represent  $\varphi(Q)$ in terms of a scalar product, $\varphi(Q)=w(x)\langle \tau_x(f), t_Q \rangle_{S'}$, for some SSP operator $S'$, which would guarantee that if there is a match between $f$ and $t_R$  at $x$, then $\varphi(Q)$ attains its global maximum (and is equal to $1$) at $Q=R$.

Let us compute $\varphi(Q)$:
\begin{eqnarray*}
\varphi(Q) &=& C_n(x)\cdot Q^{\odot n}
=  Q^{\odot n}\cdot \int_{SO(3)}R^{\odot n}c(x,R)dR \\
&=&  \int_{SO(3)}Q^{\odot n}\cdot R^{\odot n}c(x,R)dR \\
&=&  \int_{SO(3)}(Q\cdot R)^{n}c(x,R)dR \\
\end{eqnarray*}
\begin{eqnarray*}
&=&  \int_{SO(3)}(\mathbf{Re} (R^{-1}Q))^{n}c(x,R)dR \text{  (using \eqref{producto2})}\\
&=&  \int_{SO(3)}(\mathbf{Re} (R^{-1}Q))^{n}w(x) \langle \tau_x(f),t_R\rangle_S dR 
\end{eqnarray*}
(by definition of $c(x,R)$). Hence, dividing by $w(x)$, we get:
\begin{eqnarray*}
&\ & \frac{1}{w(x)}\varphi(Q)\\
&=&  \int_{SO(3)}(\mathbf{Re} (R^{-1}Q))^{n}\int_{\mathbb{R}^3} f(z+x)(O_{R^{-1}}\circ S)(t)(z) dz dR\\
&=&   \int_{SO(3)}\int_{\mathbb{R}^3} f(z+x)(O_{R^{-1}}\circ S)(t)(z) (\mathbf{Re} (R^{-1}Q))^{n}dz dR\\
&=&  \int_{\mathbb{R}^3} f(z+x) \int_{SO(3)}(O_{R^{-1}}\circ S)(t)(z) (\mathbf{Re} (R^{-1}Q))^{n}dR dz\\
&=&  \int_{\mathbb{R}^3} f(z+x) \left(\int_{SO(3)}S(t)(z)(R) (\mathbf{Re} (R^{-1}Q))^{n}dR \right)dz 
\end{eqnarray*}
(where $S(t)(z)(R)=(O_{R^{-1}}\circ S)(t)(z)$)
\begin{eqnarray*}
&=&   \int_{\mathbb{R}^3} f(z+x) (S(t)(z)\circledast_{SO(3)} K)(Q) dz 
\end{eqnarray*}
where  $K(R)=(\mathbf{Re } (R))^n$  and $$(a\circledast b)(Q)=\int_{ SO(3)} a(R)b(R^{-1}Q)dR $$ denotes the convolution of functions $a,b\in L^2(SO(3))$.

In other words, 
\begin{equation}
\varphi(Q)=Q^{\odot n} \cdot C_n(x) = w(x)\langle \tau_x(f), (S(t)\circledast_{SO(3)} K)(Q)\rangle  
\end{equation}
Here, $$(S(t)\circledast_{SO(3)} K)(Q)=\int_{SO(3)}S(t)_R K(R^{-1}Q)dR$$ must be interpreted as a function defined on $\mathbb{R}^3$ with values on $\mathbb{R}$ (indeed, it is an element of $L^2(\mathbb{R}^3)$): 
\begin{eqnarray*}
 (S(t)\circledast_{SO(3)} K)(Q)(z) &=&  \left(\int_{SO(3)}S(t)_R K(R^{-1}Q)dR\right)(z)\\ &:=& \int_{SO(3)}S(t)_R(z) K(R^{-1}Q)dR \\
&=&   (S(t)(z)\circledast_{SO(3)} K)(Q),
\end{eqnarray*}
where $S(t)(z):SO(3)\to\mathbb{R}$ is given by 
\[
S(t)(z)(R)=S(t)_R(z)=(O_{R^{-1}}\circ S)(t)(z)=S(t)(R^{-1}z).
\]
    
Indeed, in general, every element $t\in L^2(\mathbb{R}^3)$ can be interpreted, for each $z\in\mathbb{R}^3$, as an element of $L^2(SO(3))$  just making $t(z)(R)=t_R(z)$. Consequently, 
$(t\circledast_{SO(3)} K)(I_d)$ is an element of $L^2(\mathbb{R}^3)$, 
\begin{equation}
    (t\circledast_{SO(3)} K)(I_d)(z)=(t(z)\circledast_{SO(3)} K)(I_d).
\end{equation}

It follows that 
\begin{eqnarray*}
 (S(t)\circledast_{SO(3)} K)(Q)(z)
&=& \int_{SO(3)}S(t)_R(z) K(R^{-1}Q)dR\\
&=& \int_{SO(3)}S(t)_{QP}(z) K(P^{-1})dP 
\end{eqnarray*}

(in the last equality, set $R=QP$ and use that  $|Q|=1$)
\begin{eqnarray*}
&=& \int_{SO(3)}(O_{{(QP})^{-1}}\circ S)(t)(z) K(P^{-1})dP \\
 &=& \int_{SO(3)}(O_{P^{-1}Q^{-1}}\circ S)(t)(z) K(P^{-1})dP \\
 &=& \int_{SO(3)}(O_{Q^{-1}}\circ O_{P^{-1}}\circ S)(t)(z) K(P^{-1})dP \\
 &=& \int_{SO(3)}O_{Q^{-1}}(S(t)_P)(z) K(P^{-1})dP\\
  &=& O_{Q^{-1}}\left(\int_{SO(3)}S(t)_PK(P^{-1})dP\right)(z)\\
 &=& \left(\int_{SO(3)}S(t)_P K(P^{-1})dP\right)_Q(z) \\
 &=& (S(t)\circledast_{SO(3)} K)(I_d)_Q(z),
\end{eqnarray*}
where $I_d$ denotes the identity rotation. 

Let us now denote by $S_2:L^2(\mathbb{R}^3)\to L^2(\mathbb{R}^3)$ the operator given by $$S_2(t)=(t\circledast_{SO(3)} K)(I_d)$$
and let $S'=S_2\circ S$. 
Then 
\begin{eqnarray*} Q^{\odot n} \cdot C_n(x) &=& w(x)\langle \tau_x(f), (S(t)\circledast_{SO(3)} K)(Q)\rangle \\
&=&  w(x)\langle \tau_x(f), (S(t)\circledast_{SO(3)} K)(I_d)_Q\rangle \\
&=&  w(x)\langle \tau_x(f), S_2(S(t))_Q \rangle \\
&=&  w(x)\langle \tau_x(f), S_2(S(t_Q)) \rangle 
\end{eqnarray*}
(since $S, S_2$ commute with rotations)
\begin{eqnarray*}
&=&  w(x)\langle \tau_x(f), S'(t_Q) \rangle \\
&=&  w(x)\langle \tau_x(f), t_Q \rangle_{S'}  
\end{eqnarray*}

Thus, the proof ends as soon as we demonstrate that $S'$ is an SSP operator, and it is for this that we need to use Lemma \ref{composition}. Indeed, $S'=S_2\circ S$ is a composition of operators, $S$ is, by hypothesis, symmetric semidefinite positive, and  $S $, $S_2$ commute because $S$ commutes with rotations and $S_2$ is defined in terms of convolution in $SO(3)$.  Thus, Lemma \ref{composition} implies that $S'$ is SSP  whenever $S_2$ is SSP.

To prove that $S_2$ is symmetric semidefinite positive, we use the properties of the convolution on $SO(3)$ when interpreted as a hyperspherical convolution on $S^3$, the unit sphere of $\mathbb{R}^4$. Recall that if $S^{d-1}=\{x\in\mathbb{R}^d: x\cdot x^t=1\}$ denotes the (unit) sphere of $\mathbb{R}^d$, then $SO(d)$ acts transitively on $S^{d-1}$ (which means that, given $z_1,z_2\in S^{d-1}$ there is a rotation $R\in SO(d)$ such that $R(z_1)=z_2$), which makes of $S^{d-1}$ a homogeneous space and allows to introduce the convolution of functions defined on $S^{d-1}$ as follows:
\[
(f\ast_{S^{d-1}} g)(z)=\int_{SO(d)}f(R\eta)g(R^{-1}z)dR,
\]
where $\eta\in S^{d-1}$ is the north pole of the sphere and $f,g\in L^2(S^{d-1})$. 
Thus, if we use that the elements of $SO(3)$ are parametrized by quaternions of norm $1$, which that can be identified with the elements of the sphere $S^3=\{x\in \mathbb{H}: x\overline{x}=|x|=1\}$ in fourth-dimensional space, then assuming that the north pole of $S^3$ is given precisely by the identity rotation $I_d$, the convolution of $f,g\in L^2(SO(3))$ can be interpreted as a hyperspherical convolution on $S^3$:
\begin{equation}
\begin{split}
    (f\circledast_{SO(3)} g)(Q) = \int_{SO(3)}f(RI_d)g(R^{-1}Q)dR \\
    = (f\ast_{S^3} g)(Q).
\end{split}
\end{equation}
Now, as it is well known, $L^2(S^3)$ is a Hilbert space and the so-called hyperspherical harmonics,
$\{\Xi_{M}^\ell\}$, form an orthonormal basis of this space. Thus, every function $f\in L^2(S^3)$ admits a Fourier expansion
\begin{eqnarray}
    f(z)&=&\sum_{\ell, M}\hat{f}(\ell,M) \Xi_{M}^\ell(z)\\
    \hat{f}(\ell,M) &=& \langle f,\Xi_{M}^\ell(z)\rangle = \int_{S^{3}}f(\xi) \overline{\Xi_{M}^\ell(\xi)}d\xi.
\end{eqnarray}
Moreover, in \citep{Dokmanic2009}, it was proven that, if $f,g\in L^2(S^3)$ and $\frak{f}=f\ast_{S^3} g$, then 
\begin{equation}
    \hat{\frak{f}}(\ell,M)= (\ell +1)\hat{f}(\ell,M)\hat{g}(\ell,0)
\end{equation}
It follows that, given a template $t$, for each $z\in\mathbb{R}^3$, the map $t(z)(R)=t_R(z)$ belongs to $L^2(S^3)$ (here the rotations $R$ are parametrized as unit quaternions, so that $R\in S^3$) and 
\begin{eqnarray}
    t(z)(R) &=&\sum_{\ell, M}\widehat{t(z)}(\ell,M) \Xi_{M}^\ell(R)\\
    K(R) &=& \sum_{\ell, M}\widehat{K}(\ell,M) \Xi_{M}^\ell(R),
\end{eqnarray}
and 
\begin{equation}
          (t(z)\circledast_{SO(3)} K)(R) 
       =  \sum_{\ell, M} \widehat{t(z)}(\ell,M) \widehat{K}(\ell,O) (\ell +1)\Xi_{M}^\ell(R)
\end{equation}
We need the following Lemma, whose proof is included in section \ref{seccionlema}: 
\begin{lemma} \label{coeficientesK}
 $\widehat{K}(\ell,O)\geq 0$ for all $\ell$.
\end{lemma}
Then
\begin{equation}
           \langle t(z), (t(z)\circledast_{SO(3)} K)(R)\rangle_{L^2(SO(3))} 
           =  \sum_{\ell, M} (\widehat{t(z)}(\ell,M))^2 \widehat{K}(\ell,O) (\ell +1) \geq 0
\end{equation}
This means that convolution with $K$, which is an operator $C_K:L^2(SO(3))\to L^2(SO(3))$, $C_K(f)=f\circledast_{SO(3)}K$,  is semidefinite positive. Moreover, it is well known that this operator is symmetric (and we will use both things in our computations bellow).

In order to prove that $S_2$ is SSP, we introduce the operator $L:L^2(\mathbb{R}^3)\to \mathcal{C}(SO(3), L^2(\mathbb{R}^3))$ defined by $L(t)(R)=t_R$, as well as the operator
$L^*:  \mathcal{C}(SO(3), L^2(\mathbb{R}^3))\to L^2(\mathbb{R}^3)$ defined by $L^*(a)=\int_{SO(3)}a(R)_{R^{-1}}dR$. 

Then 
\begin{eqnarray*}
 \langle f,L^*(a)\rangle &=& \int_{\mathbb{R}^3}f(z)\left(\int_{SO(3)}a(R)_{R^{-1}}(z)dR\right)dz \\
&=& \int_{\mathbb{R}^3}\int_{SO(3)}f(z)a(R)(Rz)dRdz \\
&=& \int_{SO(3)}\int_{\mathbb{R}^3}f(z)a(R)(Rz)dzdR \\
&=& \int_{SO(3)}\int_{\mathbb{R}^3}f(R^{-1}w)a(R)(w)dwdR \quad \text{(just take  } w=Rz \text{ )} \\ 
&=& \int_{\mathbb{R}^3}\int_{SO(3)}f(R^{-1}w)a(R)(w)dRdw \\
&=& \int_{\mathbb{R}^3}\left(\int_{SO(3)}L(f)(R)(w)a(R)(w)dR\right)dw \\
&=& \int_{\mathbb{R}^3}\langle L(f)(w),a(w)\rangle_{SO(3)}dw,\\
\end{eqnarray*}
where  $L(f)(w)(R):=L(f)(R)(w)=f_R(w)=f(R^{-1}w)$ and $a(w)(R):=a(R)(w)$.  
On the other hand, 
\begin{eqnarray*} 
S_2(t) &=& (t\circledast_{SO(3)} K)(I_d) \\
&=& \int_{SO(3)}t_R K(R^{-1})dR \\
&=& \int_{SO(3)}L(t)(R)K(R^{-1})dR \\
&=& (L(t)\circledast_{SO(3)} K)(I_d)
\end{eqnarray*} 
Thus, if $V=\int_{SO(3)}dR$ is the volume of $SO(3)$, then
\begin{eqnarray*} 
V S_2(t) &=& (L(t)\circledast_{SO(3)} K)(I_d) \int_{SO(3)}dR \\
&=& \int_{SO(3)}(L(t)\circledast_{SO(3)} K)(I_d) dR \\
&=& \int_{SO(3)}(L(t)\circledast_{SO(3)} K)(I_d)_{R^{-1}R} dR \\
&=& \int_{SO(3)}(L(t)\circledast_{SO(3)} K)(R)_{R^{-1}} dR \\
&=& L^*(L(t)\circledast_{SO(3)} K),
\end{eqnarray*} 
where we have used that $(L(t)\circledast_{SO(3)}K)(I_d)_{R^{-1}R}=((L(t)\circledast_{SO(3)}K)(I_d)_R)_{R^{-1}}$, and that 
\begin{eqnarray*}
 (L(t)\circledast_{SO(3)}K)(I_d)(z) 
&=& (L(t)(z)\circledast_{SO(3)}K)(I_d)\\
&=& \int_{SO(3)}L(t)(z)(Q)K(Q^{-1}I_d)dQ\\
&=& \int_{SO(3)}t_Q(z)K(Q^{-1}I_d)dQ\\
&=& \int_{SO(3)}t(Q^{-1}z)K(Q^{-1}I_d)dQ
\end{eqnarray*}
so that 
\begin{eqnarray*}
 (L(t)\circledast_{SO(3)}K)(I_d)_{R}(z) 
&=& (L(t)\circledast_{SO(3)}K)(I_d)(R^{-1}z)\\
&=& \int_{SO(3)}t(Q^{-1}R^{-1}z)K(Q^{-1}I_d)dQ\\ 
&=& \int_{SO(3)}t(\Theta^{-1}z)K(\Theta^{-1} R)dQ 
\end{eqnarray*}
(set  $\Theta^{-1}=Q^{-1} R^{-1}$, so that  $Q^{-1} =\Theta^{-1} R$) \\ 
\begin{eqnarray*}
&=& \int_{SO(3)}L(t)(z)(\Theta)K(\Theta^{-1} R)dQ \\
&=& (L(t)(z)\circledast_{SO(3)}K)(R) \\
&=& (L(t)\circledast_{SO(3)}K)(R)(z).
\end{eqnarray*}

It follows that 
\begin{eqnarray*} 
\langle t,S_2(t)\rangle 
&=& \frac{1}{V} \langle t, L^*(L(t)\circledast_{SO(3)} K)\rangle  \\
&=& \frac{1}{V} \int_{\mathbb{R}^3}\langle L(t)(w),(L(t)\circledast_{SO(3)} K)(w)\rangle_{SO(3)}dw\\
&=& \frac{1}{V} \int_{\mathbb{R}^3}\langle L(t)(w),(L(t)(w)\circledast_{SO(3)} K)\rangle_{SO(3)}dw \geq 0.
\end{eqnarray*} 
Thus, $S_2(t)$ is semidefinite positive. 

Moreover, the same type of computation shows that  
\begin{eqnarray*}
 \langle f,S_2(g)\rangle 
&=&  \frac{1}{V} \langle f, L^*(L(g)\circledast_{SO(3)} K)\rangle  \\
&=& \frac{1}{V} \int_{\mathbb{R}^3}\langle L(f)(w),(L(g)(w)\circledast_{SO(3)} K)\rangle_{SO(3)}dw\\
&=& \frac{1}{V} \int_{\mathbb{R}^3}\langle (L(f)(w)\circledast_{SO(3)} K),L(g)(w)\rangle_{SO(3)}dw\\
&=& \frac{1}{V} \int_{\mathbb{R}^3}\langle L(g)(w),(L(f)(w)\circledast_{SO(3)} K)\rangle_{SO(3)}dw\\
&=& \frac{1}{V} \langle g, L^*(L(f)\circledast_{SO(3)} K)\rangle  \\
&=& \langle g,S_2(f)\rangle,  
\end{eqnarray*}  
which proves that $S_2$ is symmetric. 
This ends the proof of the theorem. 
{\hfill $\Box$} 

Note that Theorem \ref{main} connects the problem of finding, at a given position $x$, the rotation $R$ which gives a match between $f$ and $t_R$ at $x$ with the problem of finding the dominant $Z$-eigenvalue-eigenvector pair by solving \eqref{Zeigen} with $A=C_n(x)\in S^n(\mathbb{R}^4)$ and $n$ even.

\subsection{Finding the correct position} \label{position}

Although we can find the spatial positions of peaks
by running an algorithm to find the dominant $Z$-eigenvalue-eigenvector
pair for each and every voxel, this is fairly
expensive using the current decomposition algorithms for
higher degree tensors. However, the Frobenius norm of a
tensor is related to its spectral norm, and in practice it
turns out it can be used as an excellent proxy for finding the
spatial locations of peaks. Indeed, we know that $C_n(x)\in S^n(\mathbb{R}^{d'})=S^n(\mathbb{R}^{4})$. Now, if $\|T\|_{\sigma}$ denotes the  spectral norm of tensor $T$ and $\|T\|_F$ denotes its Frobenius norm, it is well-known that the largest singular value of $T$ equals its spectral norm, and that 
\[
\|T\|_{\sigma}\geq \|T\|_F\frac{1}{\sqrt{4^{n-1}}}= \|T\|_F\frac{1}{2^{n-1}}
\]
(see e.g., \citep{CaoExtreme, Tonelli2022}). 

    In fact, the connection between $\|T\|_{\sigma}$ and $\|T\|_F$ is stronger than just this inequality. As is well-known, every tensor is a finite sum of tensors of rank $1$ (indeed, if the tensor is symmetric, the tensors of rank one can also be chosen symmetric) \citep{Comon}. Moreover, if $W_1$ is a tensor of rank $1$ satisfying 
\begin{equation*}
    \|T-W_1\|_F=E_1(T):=\min_{\text{rank}(W)=1}\|T-W\|_F
\end{equation*}
then (see, e.g., \citep{RegaliaKofidisHOPM})
\begin{equation*} 
    \|W_1\|_F=\|T\|_{\sigma} 
\end{equation*}
and 
\begin{equation*} 
     E_1(T)^2= \|T\|_F^2-\|T\|_{\sigma}^2 
\end{equation*}
Thus, 
\begin{equation*} 
 \|T\|_F^2=\|T\|_{\sigma}^2+E_1(T)^2 .
\end{equation*}
Hence if $E_1(T)$ is preserved, an increment on the size of $\|T\|_{\sigma}$ ($\|T\|_F$, respectively) is translated into an increment on the size of $\|T\|_F$ ($\|T\|_{\sigma}$, respectively).

Moreover, in 1938 in \citep{Banach1938} was demonstrated that, for any symmetric tensor $T$, 
\begin{equation*} 
\|T\|_{\sigma}=\max_{\|Q\|=1}\left|\langle T,Q^{\odot n}\rangle \right|= \max_{\|Q\|=1}\left| T\cdot Q^{\odot n} \right|
\end{equation*}
Thus, large $\|T\|_F$ implies large spectral norm of $T$, and the spectral norm of $C_n(x)$ is strongly connected to the optimization problem solved in Theorem \ref{main}, which justifies using the Frobenius norm of $C_n(x)$ as a parameter to select positions $x$ where a match is possible.  

For each position $x$ identified as a
potential peak, the SS-HOPM algorithm (see \citep{Kofidis2001OnTB, Kolda2010ShiftedPM, RegaliaKofidisHOPM} for precise definition and implementation of this algorithm) is used to find the exact
dominant $Z$-eigenvalue and its associated $Z$-eigenvector, which is the rotation $R$ candidate to give a match at $x$. 

We have just explained an heuristics to locate the positions -and, after that, the rotations- where a match is possible. Now, sometimes a false positive may occur. Indeed, in the previous subsection we showed that the tensor-based
correlation function $\langle C_n(x),Q^{\odot n}\rangle$  can be seen as using a slightly different
degenerate inner product, based on $S'$  from the proof of
Theorem \ref{main}, rather than $S$. Concretely, we proved that 
\[
\langle C_n(x),Q^{\odot n}\rangle = w(x)\langle \tau_x(f), t_Q \rangle_{S'}  
\]
where $w(x)=\frac{1}{\|P_S(\tau_x(f))\|_S}$, $\|t_R\|_S=1$ and $t_R=P_S(t_R)$. This implies that the relation $-1\leq \langle C_n(x),Q^{\odot n}\rangle\leq 1$ does not necessarily hold because the normalizations were taken in terms of $S$ instead of $S'$. Taking $S'$ into account would lead to the equality
\[
\langle C_n(x),Q^{\odot n}\rangle = \frac{\langle \tau_x(f), t_Q \rangle_{S'} }{\|\tau_x(f)\|_{S'}\|t_Q\|_{S'}} w(x) \|\tau_x(f)\|_{S'}\|t_Q\|_{S'}
\]
where $-1\leq \frac{\langle \tau_x(f), t_Q \rangle_{S'} }{\|\tau_x(f)\|_{S'}\|t_Q\|_{S'}}\leq 1$ (and it is equal to $1$ when we have a match). 

So what is the impact of this? First
of all, observe that the operation that is missing from $S$ in
the normalization is effectively a kind of convolution, so that
its effect on the constant component of an
image is to scale it. Consequently, if an image is $S$-orthogonal to $\mathbf{1}$, it will
also be $S'$-orthogonal to $\mathbf{1}$. However, the norms are affected.

For the template $t$, this means that the normalization is off by
a certain factor, but this factor is the same everywhere. For the image $f$, the impact is less benign though,
as $\|\tau_x(f)\|_S$ will differ from $\|\tau_x(f)\|_{S'}$ 
in a nonuniform way.

When will this shortcoming would cause a false positive? For
this to happen, the normalization factor used at a non-match
position would have to be much higher than the ``correct''
normalization factor, and/or the normalization factor would
have to be too low at a match position. Since the
difference between $S$ and $S'$ is essentially a smoothing
operation, and the normalization factor is the reciprocal of the
norm of the projected image, the image would thus have to
be (very) smooth at the non-match position, while exhibiting
a lot of high frequency energy around the match position.
Such a situation would not be impossible, but would at the
very least be unusual in the context of a typical application
like the analysis of electron microscopy images.

\section{Proof of Lemma \ref{coeficientesK}} \label{seccionlema}
Let us start recalling the formulae associated to Fourier expansions in hyperspherical harmonics on the sphere $S^3$. The parametrization of the sphere we consider is the following one:
\[
\left\{\begin{array}{llll}
    a & =& \cos \theta  \\
    b & =& \sin \theta \cos \phi \\
    c & =& \sin \theta \sin \phi \cos \varphi \\
    d & =& \sin \theta \sin \phi \sin \varphi \\
\end{array}\right.
\ \ \text{ with }  \quad 
\left\{\begin{array}{llll}
    0 \leq \theta \leq \pi    \\
    0 \leq \phi \leq \pi    \\
    0 \leq \varphi < 2\pi    \\
\end{array}\right.
\]
where $(a,b,c,d)\in S^3$ is identified with the unit quaternion $Q=a+b\mathbf{i}+c\mathbf{j}+d\mathbf{k}$, which represents a rotation of three dimensional euclidean space $\mathbb{R}^3$. The volume element (used for integration on $S^3$ and, henceforth, also in $SO(3)$) is then given by 
\[
dV=\sin^2\theta \sin\phi d\theta d\phi d\varphi.
\]
Then every function $f\in L^2(S^3)$ can be decomposed as 
\[
f(\theta,\phi,\varphi)=\sum_{\ell=0}^{\infty} \sum_{k_2=-\ell}^{\ell} \sum_{k_1=|k_2|}^{\ell} \hat{f}(\ell,(k_1,k_2))\Xi_{(k_1,k_2)}^{\ell} (\theta,\phi,\varphi),
\]
where $\{\Xi_{(k_1,k_2)}^{\ell}\}$ denotes the orthonormal basis of $L^2(S^3)$ formed by the hyperspherical harmonics and  $\hat{f}(\ell,(k_1,k_2))=\langle f, \Xi_{(k_1,k_2)}^{\ell}\rangle_{S^3}$ are the Fourier coefficients of $f$ in this basis. We want to prove that $\hat{K}(\ell,(0,0))\geq 0$ for all $\ell$. 
Now,  $K(Q)=(\text{Re}(Q))^n=a^n=(\cos \theta)^n$ and $$\Xi_{(0,0)}^{\ell} = A_{(0,0)}^{\ell} C_{\ell}^{1}(\cos \theta),$$ where $A_{(0,0)}^{\ell}$ is a positive constant and $C_{\ell}^{\lambda}(t)$ denotes the Gegenbauer polynomial of degree $\ell$, which appears as the $\ell$-th Taylor coefficient in the expansion:
$(1-2tz+z^2)^{-1}= \sum_{\ell=0}^{\infty} C_{\ell}^{1}(t)z^{\ell}$. It is well-known that  $C_{\ell}^{1}(t)=U_{\ell}(t)$ (the $\ell$-th Chebyshev's polynomial of second kind) and that $U_{\ell}(\cos\theta)=\frac{\sin ((\ell +1) \theta)}{\sin\theta}$. Hence 
$\Xi_{(0,0)}^{\ell} = A_{(0,0)}^{\ell} \frac{\sin ((\ell +1) \theta)}{\sin\theta}$  and
\begin{eqnarray*}
     \hat{K}(\ell,(0,0)) 
    &=&  A_{(0,0)}^{\ell} \langle (\cos (\theta))^n, \frac{\sin ((\ell +1) \theta)}{\sin\theta}\rangle_{S^3} \\
    &=&  A_{(0,0)}^{\ell} \int_{0}^{\pi} \int_{0}^{\pi} \int_{0}^{2\pi} (\cos (\theta))^n \frac{\sin ((\ell +1) \theta)}{\sin\theta} \sin^2\theta \sin\phi \\ 
    &\ &  \quad \quad \quad \quad   \quad \quad \quad \quad \times d\theta d\phi d\varphi \\
    &=&  A_{(0,0)}^{\ell} \left(\int_{0}^{\pi}  (\cos (\theta))^n \frac{\sin ((\ell +1) \theta)}{\sin\theta} \sin^2\theta d\theta\right) \\
     &\ &  \quad \times \left(\int_{0}^{\pi} \sin\phi  d\phi\right) \left(\int_{0}^{2\pi} d\varphi \right)\\
    &=& 4\pi A_{(0,0)}^{\ell} \int_{0}^{\pi}  (\cos (\theta))^n \frac{\sin ((\ell +1) \theta)}{\sin\theta} \sin^2\theta d\theta\\
    &=& 4\pi A_{(0,0)}^{\ell} \int_{0}^{\pi}  (\cos (\theta))^n \sin ((\ell +1) \theta)\sin\theta d\theta
\end{eqnarray*}
To estimate the integral above, we need to use a few trigonometric formulas, as well as the hypothesis that $n$ is even. Concretely, $n$ even implies that $n/2$ is an integer and $(\cos (\theta))^n=(\cos (\pi-\theta))^n$. Moreover, for $\ell$ odd, we have that $$\sin ((\ell +1) \theta)\sin(\theta)=-\sin ((\ell +1) (\pi-\theta))\sin(\pi-\theta)$$
This makes the integral equal to $0$ for $\ell\in 2\mathbb{N}+1$. 

Assume $\ell\in 2\mathbb{N}$. Then  
\begin{eqnarray*}
(\cos\theta)^n &=& \left(\frac{e^{i\theta}+e^{-i\theta}}{2}\right)^n=\frac{1}{2^n}\sum_{k=0}^n\binom{n}{k}e^{i\theta(n-k)}e^{-i\theta k} \\
&=&\frac{1}{2^n}\sum_{k=0}^n\binom{n}{k}e^{i\theta(n-2k)}\\
&=& \frac{1}{2^n}\left[ \sum_{s=0}^{n/2}\binom{n}{\frac{n}{2}-s} e^{i\theta(n-2(\frac{n}{2}-s))}  + \sum_{s=1}^{n/2}\binom{n}{\frac{n}{2}+s} e^{i\theta(n-2(\frac{n}{2}+s))}
\right]\\
\end{eqnarray*}
\begin{eqnarray*}
&=& \frac{1}{2^n}\left[ \binom{n}{\frac{n}{2}}+\sum_{s=1}^{n/2}\binom{n}{\frac{n}{2}-s} e^{2i\theta s} 
+ \sum_{s=1}^{n/2}\binom{n}{\frac{n}{2}+s} e^{-2i\theta s} \right]\\
&=& \frac{1}{2^n}\left[ \binom{n}{\frac{n}{2}}+2 \sum_{s=1}^{n/2}\binom{n}{\frac{n}{2}-s} \frac{e^{2i\theta s} + e^{-2i\theta s}}{2} 
\right]\\
&=& \frac{1}{2^n}\left[ \binom{n}{\frac{n}{2}}+2 \sum_{s=1}^{n/2}\binom{n}{\frac{n}{2}-s} \cos(2\theta s) 
\right]\\
&=& \frac{1}{2^n}\left[ \binom{n}{\frac{n}{2}}+2 \sum_{k=1}^{n/2}\binom{n}{k} \cos(\theta(n-2k)) 
\right] 
\end{eqnarray*}
(for the last line,  just set $k=n/2-s$). Moreover, it is well-known that
$$
\sin(\theta)\sin((\ell+1)\theta)= \frac{1}{2}(\cos(\ell\theta)-\cos((\ell+2)\theta)),
$$ so that, by a direct substitution in the formula defining  $\hat{K}(\ell,(0,0))$ we get
\begin{eqnarray*}
    \frac{\hat{K}(\ell,(0,0))}{4\pi A_{(0,0)}^{\ell}}  &=&  \int_{0}^{\pi}  \frac{1}{2}(\cos(\ell\theta)-\cos((\ell+2)\theta)) \\
   &\ & \quad \times \left(\frac{1}{2^n}\left[ \binom{n}{\frac{n}{2}}+2 \sum_{k=1}^{n/2}\binom{n}{k} \cos(\theta(n-2k)) 
\right] \right) d\theta \\
&=& \frac{1}{2^{n+1}} \int_{0}^{\pi}  (\cos(\ell\theta)-\cos((\ell+2)\theta))\\
   &\ & \quad \times \left[ \binom{n}{\frac{n}{2}}+2 \sum_{k=1}^{n/2}\binom{n}{k} \cos(\theta(n-2k)) 
\right]  d\theta .\\
\end{eqnarray*}
We can now use that $$\cos(x)\cos(y)=\frac{1}{2}(\cos(x+y)+\cos(x-y))$$ 
to claim that

\begin{eqnarray*}
  &\ & \frac{\hat{K}(\ell,(0,0))}{4\pi A_{(0,0)}^{\ell}}  
   =   \frac{1}{2^{n+1}} \int_{0}^{\pi} \binom{n}{\frac{n}{2}} (\cos(\ell\theta)-\cos((\ell+2)\theta))d\theta \\
   &\ & \quad +\frac{1}{2^{n}}\int_0^\pi \sum_{k=1}^{n/2}\binom{n}{k} (\cos(\ell\theta)-\cos((\ell+2)\theta)) \cos(\theta(n-2k)) d\theta\\
&= &   \frac{1}{2^{n+1}} \int_{0}^{\pi} \binom{n}{\frac{n}{2}} (\cos(\ell\theta)-\cos((\ell+2)\theta))d\theta \\
   &\ & \quad +\frac{1}{2^{n+1}}\int_0^\pi \sum_{k=1}^{n/2}\binom{n}{k} \\
   &\ &\quad \quad \quad \times [\cos(\theta(\ell+n-2k))  + \cos(\theta(\ell+2k-n)) \\
   &\ &\quad \quad \quad \quad \quad -\cos(\theta(\ell +2+n-2k))-\cos(\theta(\ell+2+2k-n)] d\theta \\
\end{eqnarray*}
The parity of $\ell$ and $n$ implies that all factors that appear multiplying the variable $\theta$ inside of the cosine functions are even numbers. This makes the corresponding integrals (on $[0,\pi]$) equal to $0$, except in the case that the factor itself is $0$. In such case,  $\cos(0)=1$ implies that only the cosine functions that appear with a minus sign in front of them in the formula can contribute with a negative number to the integral. Now clearly $\ell +2>0$ always since $\ell\geq 0$, and $\ell +2+n-2k=0$ implies $2k=n+\ell +2>n$, so that $k>n/2$ which is impossible since the sum's range goes from $k=1$ to $k=n/2$. This means that the term $-\cos(\theta(\ell +2+n-2k))$ never contributes with a negative number to the sum. On the other hand, if the cosine function with factor $\ell +2+2k-n$ contributes, which means that $\ell +2+2k-n=0$, then  $k=(n-\ell -2)/2<n/2$. In particular, taking $k^*=k+1$, we have that $1\leq k^*\leq n/2$ so that $\cos(\theta(\ell+2k^*-n))= \cos(\theta(\ell+2k+2-n))=\cos (0)=1$ and the corresponding term effectively appears in  the sum. In particular, adding these two terms of the sum we get  
\begin{eqnarray*}
   &\ & \frac{1}{2^{n+1}}\int_0^\pi [\binom{n}{k^*}\cos(\theta(\ell+2k^*-n))  -\binom{n}{k}\cos(\theta(\ell+2+2k-n)] d\theta \\
   &\ & \quad = \frac{1}{2^{n+1}}\int_0^\pi \left[\binom{n}{k+1}-\binom{n}{k}\right] d\theta \\
   &\ & \quad = \frac{\pi}{2^{n+1}}\left(\binom{n}{k+1}-\binom{n}{k}\right)>0 
\end{eqnarray*}
since  $n$  even and $k<n/2$. This ends the proof of Lemma \ref{coeficientesK}

{\hfill $\Box$}


\section{Conclusions} 

We have exposed the maths of classical template matching with rotations. Moreover, an alternative to the classical algorithm, named tensorial template matching (or TTM), has been shown. TTM integrates the information relative to all rotated versions of a template $t$ into a unique symmetric tensor template $T$, which is computed only once per template. The main theorem of the paper, Theorem \ref{main}, shows that finding an exact match between an image $f$ and a rotated version $t_R$ of the template $t$ at a given position $x$ is equivalent to finding a best rank 1 approximation (in the Frobenius norm) to a certain tensor $C_n(x)$. The resulting algorithm has reduced computational complexity when compared to the classical one. TTM finds the position and rotation of instances of the template in any tomogram with just a few correlations with the linearly independent components of $T$. In particular, Cryo-electron tomography (3D images) for macromolecular detection requires 7112, 45123 and 553680 rotations to achieve an accuracy of 13$^\circ$, 7$^\circ$ and 3$^\circ$ respectively \citep{Chaillet2023}. Therefore, and considering 4-degree tensors (35 linearly independent components), the potential speed-up of our approach with respect to TM is 203x, 1239x and 184560x in these cases, while the angular accuracy remains constant for TTM and it is limited by the computation of tensorial template. 

\section{Acknowledgements} This work is based on unpublished ideas of Jasper van de Gronde when he was a researcher of the University of Groningen. This work was supported by Ramon y Cajal program, RYC2021-032626-I, of the Spanish State Research Agency (AEI) and Attract-RYC 2023 program of the University of Murcia. 

We want to express our thanks to Holger Kohr, Erik Franken and Remco Schoenmakers from Thermo Fisher Scientific for their support and feedback about the potential of tensorial template matching.  

\bibliographystyle{plainnat}

\bibliography{mybib}


\end{document}